%% file: main_powered.tex
\documentclass[10pt,twocolumn]{article}

\usepackage[margin=0.72in,columnsep=0.24in]{geometry}
\usepackage[T1]{fontenc}
\usepackage[utf8]{inputenc}
\usepackage{lmodern}
\usepackage{microtype}
\usepackage{amsmath,amssymb}
\usepackage{booktabs}
\usepackage{tabularx}
\usepackage{caption}
\usepackage{graphicx}
\usepackage{tikz}
\usepackage{enumitem}
\usepackage{xcolor}
\usepackage{xurl}
\usepackage{seqsplit}
\usepackage{hyperref}
\usepackage[capitalise,noabbrev]{cleveref}
\usetikzlibrary{arrows.meta,positioning,fit,backgrounds}

\definecolor{FBBlue}{HTML}{1769AA}
\definecolor{FBGold}{HTML}{E6A11A}
\definecolor{FBTeal}{HTML}{168C7A}
\definecolor{FBRed}{HTML}{C75450}
\definecolor{FBCharcoal}{HTML}{262B33}
\definecolor{FBPaper}{HTML}{F7F8FA}
\hypersetup{
  colorlinks=true,
  linkcolor=FBBlue,
  citecolor=FBBlue,
  urlcolor=FBBlue,
  pdftitle={FlavourBench: Executable Culinary Reward Maps for Language Model Evaluation and Post-Training},
  pdfauthor={Josef Chen, Erim Hayretci},
  pdfsubject={An executable benchmark of culinary reasoning in frontier language models},
  pdfkeywords={language models, executable benchmarks, culinary reasoning, evaluation,
    statistical ranking}
}
\setlist{leftmargin=*,nosep}
\newcommand{\system}{\textsc{FlavourBench}}
\newcommand{\epicure}{\textsc{Epicure}}

\input{generated/complete-core/complete-core-macros.tex}

\input{generated/complete-core/complete-core-stability-macros.tex}

\input{generated/complete-core/complete-core-robustness-macros.tex}

\input{generated/complete-core/complete-core-public-scorer-macros.tex}

\input{generated/complete-core/complete-core-external-substitution-validation-macros.tex}

\input{generated/complete-core/complete-core-reward-transfer-macros.tex}

\title{\vspace{-1.4em}\system{}: Executable Culinary Reward Maps for\\
Language Model Evaluation and Post-Training}
\author{
  Josef Chen\\[-0.1em]\small Independent Researcher
  \and Erim Hayretci\\[-0.1em]\small Imperial College London
}
\date{August 2026}

\begin{document}
\raggedbottom
\twocolumn[
  \begin{@twocolumnfalse}
  \maketitle
  \vspace{-1.1em}
  \begin{abstract}
  Open-ended language-model evaluation often substitutes another model or a small preference
  panel for a missing answer key. We introduce \system{}, which instead compiles dense answer maps
  from a versioned culinary environment. Each task asks for three ingredients from eight; before
  inference, \epicure{} scores all 56 legal portfolios. We evaluate \FBModels{} frontier endpoints
  on the same \FBTasks{} substitution, pairing, and constraint tasks, yielding
  \FBCompleteCells{} complete observations. Anchor-cluster bootstraps and Holm-controlled paired
  tests resolve \FBSignificantPairs{} of \FBPairs{} model contrasts. \FBTopModel{} has the largest
  point estimate (\FBTopScore{}), but corrected evidence does not identify a unique best endpoint.
  Rankings replicate across independently compiled panels and remain similar under alternative
  metrics, task filters, family weights, and three public \epicure{} checkpoints. Those checkpoints
  also rank human-observed Recipe1MSubs targets above within-food-group chance. We then run a
  preregistered, three-seed post-training study. LoRA SFT of a pinned Qwen3-0.6B checkpoint on 270
  \epicure{}-optimal answers improves its score on \FBTransferPrimaryTasks{} anchor-disjoint maps
  by \FBTransferPrimaryGain{} points over a format- and label-matched control (95\% CI
  \FBTransferPrimaryCILow{}--\FBTransferPrimaryCIHigh{}, $p=\FBTransferPrimaryP{}$), and the
  effect replicates on all \FBTransferPublicTasks{} public maps (+\FBTransferPublicGain{} points,
  95\% CI \FBTransferPublicCILow{}--\FBTransferPublicCIHigh{}). On the primary split, both trained
  arms parse every response while format training alone does not improve on the base. The release
  contains prompts, reward maps, raw responses, routes, training and evaluation manifests, code,
  and offline verifiers.
  \end{abstract}
  \vspace{0.5em}
  \end{@twocolumnfalse}
]

\section{Introduction}

Benchmarks are strongest when the evaluator can be executed. Code can be tested, database states
can be inspected, and games can be scored. Culinary reasoning is usually evaluated with a model
judge, a handful of human preferences, or factual questions whose answers do not capture the
quality of a complete decision. That leaves two problems: the judge is entangled with the systems
being tested, and exact match throws away useful differences between plausible answers.

\system{} turns a culinary runtime into a benchmark environment. \epicure{} represents 1,790
ingredients in a 300-dimensional space and exposes deterministic operations for substitution,
pairing, dietary feasibility, and regional composition
\cite{radzikowski2026structure,radzikowski2026geometry}. A task compiler converts these operations
into three-of-eight selection problems. It enumerates every candidate portfolio and freezes a
continuous 0--100 score map before model execution. An answer is therefore scored by table lookup,
without a model judge or interpretation after the fact.

\epicure{} is not a leaderboard entry and is not ``100\% accurate.'' It defines the task-specific
reward surface. A model receives 100 on a task only if it selects that task's highest-scoring
portfolio; other valid portfolios receive partial credit. The benchmark measures agreement with a
released culinary reward map, not universal human taste.

We use that environment to evaluate \FBModels{} current endpoints. Every model faces the same
\FBTasks{} tasks and \FBSelectionsPerTask{} frozen answer scores per task. The complete common core
supports direct paired comparisons on \FBTasks{} shared observations per model pair instead of
inferring a ranking from sparse votes or comparing endpoints on different task subsets.

The paper makes four contributions:
\begin{enumerate}[label=\arabic*.]
  \item an executable culinary benchmark with \FBPrefrozenScores{} model-independent portfolio
  scores and a single interpretable 0--100 metric;
  \item a \FBModels{}-model, \FBTasks{}-task complete-core evaluation with simultaneous
  uncertainty and paired, multiplicity-controlled tests;
  \item post-collection tests of task-filter, metric, family-weight, and reward-map dependence,
  including a rescore against three public \epicure{} checkpoints and a held-out, human-observed
  substitution check; and
  \item a preregistered three-seed study showing that \epicure{}-optimal SFT transfers to unseen
  reward maps beyond format learning, backed by a public-map replication and content-addressed
  release.
\end{enumerate}

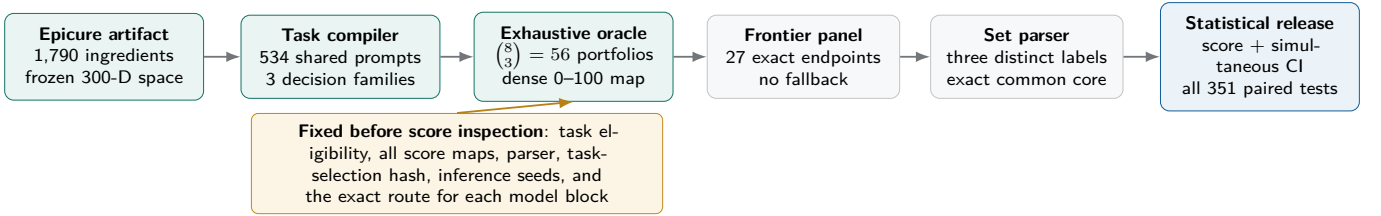
\begin{figure*}[t]
  \centering
  \resizebox{\linewidth}{!}{%
  \begin{tikzpicture}[
    x=1cm,y=1cm,font=\sffamily\footnotesize,
    box/.style={draw=FBCharcoal!28,rounded corners=4pt,fill=FBPaper,
      minimum height=1.25cm,text width=2.55cm,align=center,inner sep=6pt},
    oracle/.style={draw=FBTeal!82!black,rounded corners=4pt,fill=FBTeal!9,
      minimum height=1.25cm,text width=2.65cm,align=center,inner sep=6pt},
    score/.style={draw=FBBlue!88!black,rounded corners=4pt,fill=FBBlue!8,
      minimum height=1.25cm,text width=2.65cm,align=center,inner sep=6pt},
    arrow/.style={-{Latex[length=2.2mm]},thick,draw=FBCharcoal!65}
  ]
    \node[oracle] (runtime) at (0,0) {\textbf{Epicure artifact}\\1,790 ingredients\\frozen 300-D space};
    \node[oracle] (compiler) at (3.65,0) {\textbf{Task compiler}\\\FBTasks{} shared prompts\\3 decision families};
    \node[oracle] (enumerate) at (7.25,0) {\textbf{Exhaustive oracle}\\$\binom{8}{3}=56$ portfolios\\dense 0--100 map};
    \node[box] (models) at (10.8,0) {\textbf{Frontier panel}\\\FBModels{} exact endpoints\\no fallback};
    \node[box] (parser) at (14.25,0) {\textbf{Set parser}\\three distinct labels\\exact common core};
    \node[score] (inference) at (17.85,0) {\textbf{Statistical release}\\score + simultaneous CI\\all 351 paired tests};

    \draw[arrow] (runtime) -- (compiler);
    \draw[arrow] (compiler) -- (enumerate);
    \draw[arrow] (enumerate) -- (models);
    \draw[arrow] (models) -- (parser);
    \draw[arrow] (parser) -- (inference);

    \node[draw=FBGold!85!black,rounded corners=3pt,fill=FBGold!11,
      text width=6.0cm,align=center,inner sep=5pt] (freeze) at (5.45,-1.65)
      {\textbf{Fixed before score inspection}: task eligibility, all score maps, parser,
      task-selection hash, inference seeds, and the exact route for each model block};
    \draw[arrow,draw=FBGold!80!black] (freeze.north) -- (enumerate.south);
  \end{tikzpicture}}
  \caption{\system{} compiles a content-addressed culinary artifact into dense selection tasks. The
  evaluated models never judge one another and do not access \epicure{}. All score maps and the
  per-panel contracts are fixed before collection. The complete-core task set is selected from
  response validity alone, before inspecting any model selection or score.}
  \label{fig:architecture}
\end{figure*}

\section{Executable culinary reference environment}
\label{sec:ground-truth}

\subsection{A reference environment, not a model judge}

The primary task maps bind one exact \epicure{} data bundle and application build by hash. The data
and application bytes are public at an immutable source revision, so the released score maps can
be reconstructed exactly. The original training run, seed, and source revision for this runtime
were not recovered, and the artifact is not identified as the published Cooc, Core, or Chem
checkpoint. We therefore use it as a fixed reference function, not as independently established
culinary truth. Changing the artifact creates a new task-set identity. A separate post-hoc analysis
below replaces its reward maps with the three public, paper-linked \epicure{} checkpoints.

For each task $t$, let $\mathcal{A}_t$ contain all 56 three-item subsets of eight candidates.
\epicure{} assigns a raw utility $u_t(a)$ to every valid $a\in\mathcal{A}_t$. Invalid portfolios
under a dietary or category constraint receive zero. Valid utilities are normalized within the
task:
\begin{equation}
 s_t(a)=100\,\frac{u_t(a)-\min_{b\in\mathcal{A}_t}u_t(b)}
 {\max_{b\in\mathcal{A}_t}u_t(b)-\min_{b\in\mathcal{A}_t}u_t(b)}.
 \label{eq:task-score}
\end{equation}
Every task has a unique 100-point optimum. The exact chance baseline is not a generic percentage;
it is $|\mathcal{A}_t|^{-1}\sum_a s_t(a)$ for that frozen task.

\subsection{Three complete-core decision families}

The ranked core contains \FBTasks{} tasks: \FBTasksPerPanelFamily{} from each family in each of two
independently compiled panels, spanning \FBUniqueAnchors{} unique anchor ingredients. Tasks sharing
an anchor move together as one cluster in inference. Candidate sets are selected across validation
strata, then ordered and labeled by a task-derived hash. No evaluated model writes an item,
distractor, or score. Regional-composition tasks are released as a supplemental track but are not
in the 27-model leaderboard because they do not support an equally complete common core.

\begin{table*}[t]
  \centering
  \footnotesize
  \setlength{\tabcolsep}{4pt}
  \begin{tabularx}{\linewidth}{@{}l X X X r@{}}
    \toprule
    Family & Decision & Epicure utility & Hard constraint & Tasks \\
    \midrule
    Substitution & choose a three-item replacement portfolio for an anchor ingredient &
      $0.8$ anchor similarity $+0.2$ portfolio coherence & none & \FBTasksPerFamily{} \\
    Pairing & choose three ingredients to accompany an anchor &
      $0.65$ anchor affinity $+0.35$ portfolio coherence & none & \FBTasksPerFamily{} \\
    Constraints & choose a feasible portfolio under diet and processing limits &
      $0.7$ anchor similarity $+0.3$ coherence & diet and maximum NOVA level & \FBTasksPerFamily{} \\
    \bottomrule
  \end{tabularx}
  \caption{The three complete-core task families share one response and scoring interface but
  probe different culinary decisions. Each task contains eight candidates and 56 frozen scores.}
  \label{tab:task-families}
\end{table*}

The score maps have substantial resolution rather than one correct key plus 55 equivalent
mistakes. \Cref{tab:task-diagnostics} reports the exact random-portfolio baseline, separation of
the best and second-best portfolios, and the number of distinct score values. Constraint tasks are
intentionally sparse because infeasible portfolios score zero; the other families provide much
finer graded rewards.

\begin{table}[t]
  \centering
  \scriptsize
  \setlength{\tabcolsep}{3pt}
  \input{generated/complete-core/complete-core-task-table.tex}
  \caption{Frozen task-map diagnostics for \FBTasksPerFamily{} scheduled tasks per family,
  computed before model execution.
  Chance is the uniformly random portfolio mean; top gap is the median best--second-best margin.}
  \label{tab:task-diagnostics}
\end{table}

\subsection{Prompt and parser contract}

Prompts show the decision context and eight labeled candidates. The final line must begin with the
\texttt{FINAL\_SELECTION} marker and contain three distinct A--H labels separated by commas. No
prompt supplies a concrete valid triplet. The parser takes the final exact marker line, uppercases
the labels, sorts them as an unordered set, and performs one lookup in the frozen score map.
Explanatory text does not affect the result. Missing markers, duplicate labels, provider failures,
and abnormal finishes remain in the raw response ledger but cannot enter the common core.

The model-only condition is deliberate. The fixed reward map is not exposed as a tool during
ranking. The score therefore measures whether a model can make the culinary decision, rather than
whether it can copy a runtime return value after being told which operation to call.

\section{Evaluation design}
\label{sec:design}

\subsection{One primary score}

For model $m$, panel $p$, family $f$, and task $t$, let $Y_{mt}$ be the frozen portfolio score and
let $T_{pf}$ be the score-blindly selected set of \FBTasksPerPanelFamily{} tasks completed by every
model. The FlavourBench Score is
\begin{equation}
 F_m=\frac{1}{2\cdot3}\sum_{p=1}^{2}\sum_{f=1}^{3}
      \frac{1}{|T_{pf}|}\sum_{t\in T_{pf}}Y_{mt}.
 \label{eq:fb-score}
\end{equation}
It ranges from 0 to 100 and equals the ordinary mean over all \FBTasks{} cells because the six
panel--family strata have equal size. Every ranked endpoint contributes exactly \FBTasks{} valid
responses. Failures are neither assigned zero nor silently dropped model by model: they determine
whether a task can enter the shared core, using response status and parser validity only. The
task-selection hash is fixed before loading any selected portfolio or score.

\subsection{Frontier endpoint panel}

The panel spans OpenAI, Anthropic, Google, xAI, Meta, Kimi, Qwen, Z.ai, DeepSeek, ByteDance,
Thinking Machines, MiniMax, NVIDIA, Mistral, Tencent, and Cohere. It includes GPT-5.6 Sol, Terra,
and Luna; Claude Opus, Sonnet, and Fable 5; Gemini 3.1 Pro and 3.6 Flash; Grok 4.6; Llama 4
Maverick; Muse Spark 1.2 and Muse Glimmer 30B; Kimi K3; Qwen 3.8 Max and Qwen3.8 2.4T A95B;
GLM 5.2 and GLM 5.3; DeepSeek V4 Pro 0813 and V4 Flash; Seed 2.1 Turbo; Inkling; MiniMax M3; Nemotron 3.5
Lightning; Mistral Large 3; Tencent HY 3; Cohere Command A; and Cohere Command R+ (08-2024).
Z.ai supplied written permission for one finite GLM 5.3 benchmark run, conditioned on the adapter
not becoming a permanent running function. We bind that permission and its hash in the release,
require the returned identity to equal \texttt{glm-5.3}, and expose no standing endpoint.\footnote{\url{https://docs.z.ai/guides/llm/glm-5.3}}

Routes were chosen before each scored block and automatic fallback is disabled. Most endpoints
use an exact OpenRouter route; Qwen 3.8 Max uses Alibaba, Fable 5 uses Google Vertex, GLM 5.3 uses
the finite Z.ai route above, and selected DeepSeek blocks use an exact GMICloud or DeepInfra route.
When transport validation required a replacement, the complete model block was rerun under one
new route; successful cells from superseded blocks are not pooled. The manifest binds the requested
model, returned identity contract, endpoint tag, provider, supported parameters, and execution
contract for both panels. The exact per-model routes appear in \cref{app:routes}.
For endpoints that advertise a reasoning-effort control, hidden reasoning is excluded and effort
is fixed to the lowest transport-stable setting recorded in the exact analysis plan. Endpoints
without that control use their provider-fixed behavior. This prevents hidden deliberation from
silently consuming the answer budget while leaving the scored response format identical across
routes.

\subsection{Score-blind complete-core construction}

The two task compilers each produce 160 candidates per family. After collection, the release
re-scores response syntax with one versioned parser and constructs a validity matrix. Within every
panel--family stratum it selects the first \FBTasksPerPanelFamily{} eligible task IDs under a fixed
SHA-256 ordering. A task is eligible only when all \FBModels{} models completed it and produced a
parseable three-item set. The selection code does not load the chosen ingredients or their scores
until the task IDs are fixed. This yields a balanced common core without favoring any model or
post-selecting easy tasks by observed score.

\subsection{Uncertainty and multiplicity}

All inferential choices are bound in the content-addressed analysis plan.
\begin{itemize}
  \item \textbf{Scores.} \FBBootstrapResamples{} bootstrap samples are drawn over ingredient
  anchors; tasks that share an anchor across panels always move together. Each resample is reduced
  to six panel--family means and then equally weighted. We report pointwise
  percentile intervals and a simultaneous max-$t$ band across all \FBModels{} model scores.
  \item \textbf{Pairs.} Every one of the \FBPairs{} model pairs uses the same \FBTasks{} tasks.
  Two-sided anchor-cluster sign-flip tests use
  \FBPermutationResamples{} Monte Carlo draws and Holm correction at familywise $\alpha=.05$.
  We also report paired mean differences, bootstrap intervals, and Cohen's $d_z$.
  \item \textbf{Chance.} Each model is compared with the exact taskwise mean over all 56
  portfolios. The \FBModels{} paired sign-flip tests form a separate Holm family.
  \item \textbf{Ranks.} Point ranks are accompanied by bootstrap rank intervals and contiguous
  statistical groups. Models that are not significantly separated are not presented as a resolved
  total order.
\end{itemize}

We add a retrospective precision diagnostic that does not alter the confirmatory inference.
Within each of the six family--panel strata, we draw \FBStabilityReplicates{} score-blind subsets
without replacement at total task counts 30, 60, 90, 150, and 270, then compare each point order
with the complete 534-task order. We report rank correlation, top-five overlap, point-leader
preservation, pair-order agreement, and score error. A crossed model-by-task decomposition also
reports a descriptive generalizability coefficient for relative endpoint comparisons. It treats
the release as a crossed design; it does not turn these fixed endpoints or tasks into random
population samples.

Because complete-core eligibility depends on the full endpoint roster, we also run four
post-collection sensitivity checks. First, each endpoint is omitted in turn and the six task
strata are reselected under the original hash order. Second, each selected stratum is compared
with its 160-task candidate pool on five score-map properties, the mean score of the other 26
endpoints, and ingredient category. We compare the 42 observed differences with 20,000 random
subsets per stratum and apply Holm correction. Third, we recompute the leaderboard using
chance-adjusted gain, action percentile, and exact-optimum rate. Finally, we enumerate all
one-percentage-point family weights between 0.20 and 0.50. These diagnostics test whether one
roster, task filter, score formula, or moderate family weighting determines the reported order;
they do not replace the prespecified analysis.

We also test whether the primary ordering depends on the embedding behind the reward map. This
analysis is post hoc. It keeps all 534 prompts, candidate sets, constraints, and 14,418 observed
model selections fixed, then recomputes every 56-action map with Epicure-Cooc, Epicure-Core, and
Epicure-Chem at immutable Hugging Face revisions \cite{radzikowski2026geometry}. All ingredients
used by the benchmark occur in each public vocabulary. We compare task-map and aggregate model
orders with the primary map and use \FBPublicScorerReplicates{} panel-by-family stratified anchor
bootstraps for descriptive intervals. Because the original runtime selected the candidates, this
tests reward-map dependence conditional on the released tasks; it is not a newly compiled public-
checkpoint benchmark.

Finally, we test the substitution geometry of the public checkpoints against Recipe1MSubs
\cite{fatemi2023substitute}, whose standardized test split contains substitutions extracted from
recipe-user comments. The analysis protocol is hash-bound before checkpoint scores are inspected.
We use exact token matches only, deduplicate directed source--target pairs, and rank each observed
target by source--candidate cosine similarity. The primary comparison is deliberately strict: the
target is ranked only against ingredients in its own pre-existing food group, so broad category
recognition cannot by itself clear the null. Pair percentiles are averaged within source ingredient
and then equally across sources; 50,000 source-cluster bootstrap draws form 95\% intervals, with
Holm correction across the three checkpoint tests. A sensitivity set retains only directed pairs
absent from the Recipe1MSubs training split. The labels are external to Epicure training, but the
underlying recipes derive from Recipe1M, which is upstream of part of Epicure's corpus; this is
label-independent convergent validation, not corpus-independent validation.

The design has \FBIndependentClusters{} independent anchor clusters and \FBTasks{} paired cells
per contrast. Statistical claims are read directly from simultaneous intervals and Holm-adjusted
tests; point-rank differences without corrected support are shown as unresolved.

\section{Results}
\label{sec:results}

\subsection{A statistically resolved FlavourBench leaderboard}

\begin{figure*}[t]
  \centering
  \includegraphics[width=0.92\linewidth]{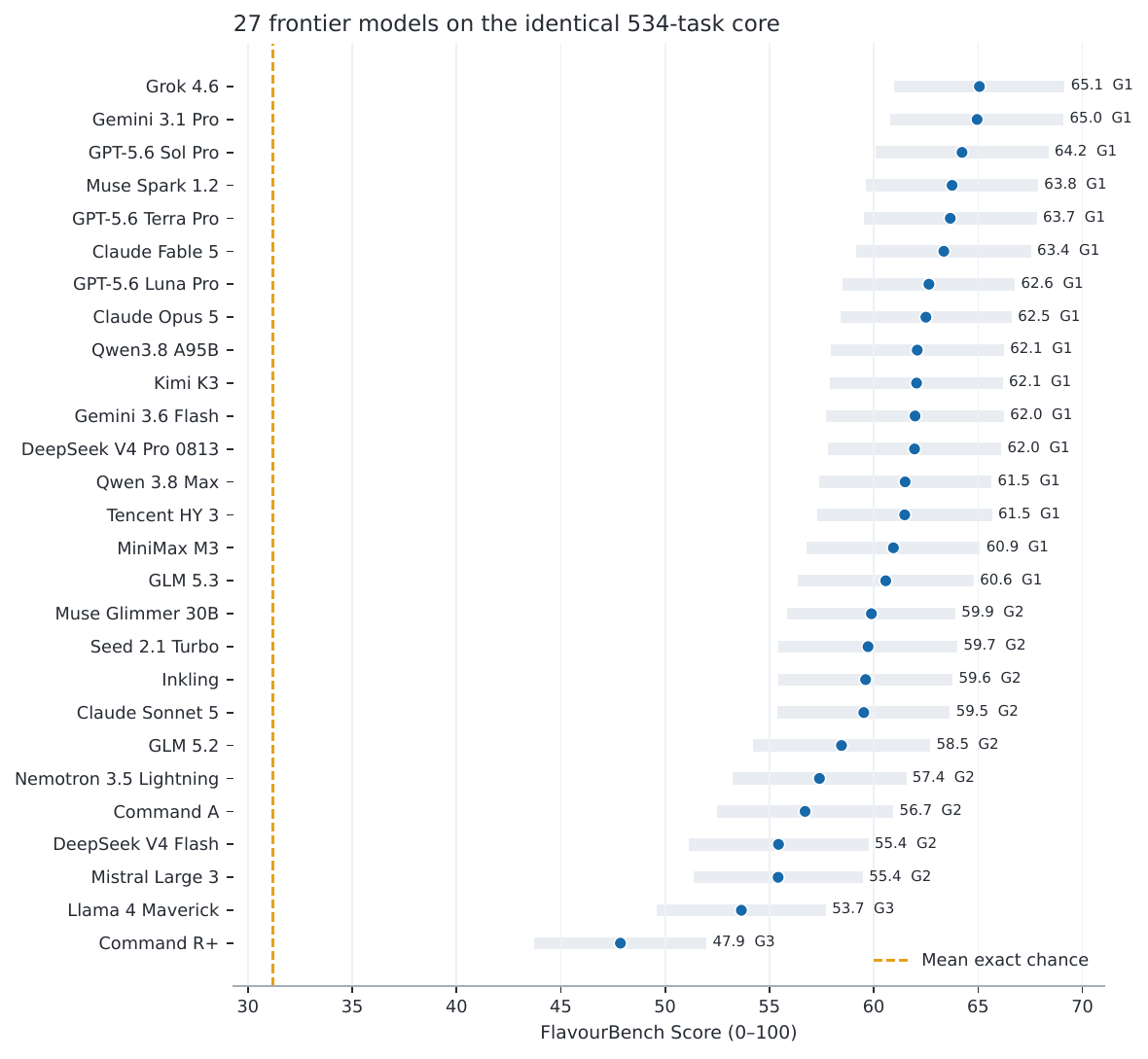}
  \caption{FlavourBench Score with simultaneous 95\% max-$t$ intervals. The letter-free group
  numbers are inferential tiers: a point rank is shown for navigation, while models inside an
  unresolved group should not be read as a statistically established ordering.}
  \label{fig:leaderboard}
\end{figure*}

All \FBModels{} endpoints receive a score on the same \FBTasks{} tasks.
\FBTopModel{} has the largest point estimate, \FBTopScore{}, with simultaneous 95\% interval
[\FBTopCILow{}, \FBTopCIHigh{}]. Across all \FBPairs{} prespecified paired contrasts,
\FBSignificantPairs{} remain significant after Holm correction. A point rank is therefore a compact
summary, not a claim that every adjacent model is distinguishable.
\cref{fig:leaderboard} shows the score scale, simultaneous intervals, and rank groups;
\cref{tab:leaderboard} gives exact values.

The ranked matrix is rectangular: every row contains \FBTasks{} valid, parser-confirmed responses,
so no endpoint is advantaged by answering a different subset. Provider failures and filtered calls
remain available in the full execution ledger, but none appears as a zero-valued culinary decision
and no incomplete row enters this leaderboard.

\begin{table*}[t]
  \centering
  \scriptsize
  \setlength{\tabcolsep}{3.6pt}
  \input{generated/complete-core/complete-core-leaderboard-table.tex}
  \caption{The complete automated leaderboard. Every score and every pairwise comparison uses the
  identical \FBTasks{}-task core. Rank intervals come from the anchor-cluster bootstrap; groups
  summarize the Holm-controlled pairwise graph.}
  \label{tab:leaderboard}
\end{table*}

\subsection{Does the ordering replicate on a second task panel?}

\begin{figure}[t]
  \centering
  \includegraphics[width=\linewidth]{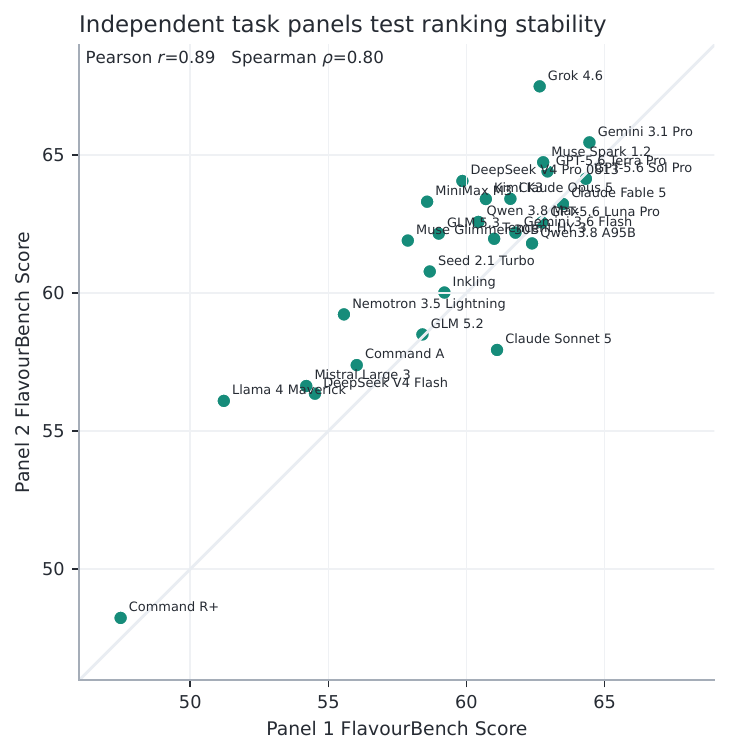}
  \caption{Panel-1 versus panel-2 FlavourBench Score for every endpoint. The diagonal marks exact
  agreement. Each point aggregates \FBTasksPerPanelFamily{} tasks per family; the joint analysis
  uses \FBUniqueAnchors{} unique anchor clusters.}
  \label{fig:panel-replication}
\end{figure}

The two panel-specific scores use the same frozen metric and \FBTasksPerPanelFamily{} tasks per
family, but disjoint task IDs.
Their model-level Pearson correlation is \FBPanelPearson{} and their rank correlation is
\FBPanelSpearman{}. These are descriptive stability diagnostics, not a model-selection rule; the
primary leaderboard and its uncertainty use both panels and cluster shared ingredient anchors.
\Cref{fig:panel-replication} exposes every model rather than reducing
replication to a single coefficient.

\subsection{How much task evidence is enough for a stable score?}

The crossed design's relative-decision generalizability coefficient is
\FBGeneralizability{} at 534 tasks; the same variance model estimates \FBTasksForGNinety{}
balanced tasks for 0.90. In \FBStabilityReplicates{} stratified half-size subsamples
(\FBHalfTaskCount{} tasks), the median rank correlation with the complete point order is
\FBHalfRankMedian{} (empirical 95\% range \FBHalfRankLow{}--\FBHalfRankHigh{}) and median
top-five overlap is \FBHalfTopFive{}. The complete point leader is retained in only
\FBHalfTopOne{} of those subsets. The first two results show that 534 tasks support a reliable
global score; the last shows why a narrow claim about the single point leader would be stronger
than the data allow.

\begin{figure*}[t]
  \centering
  \includegraphics[width=0.92\linewidth]{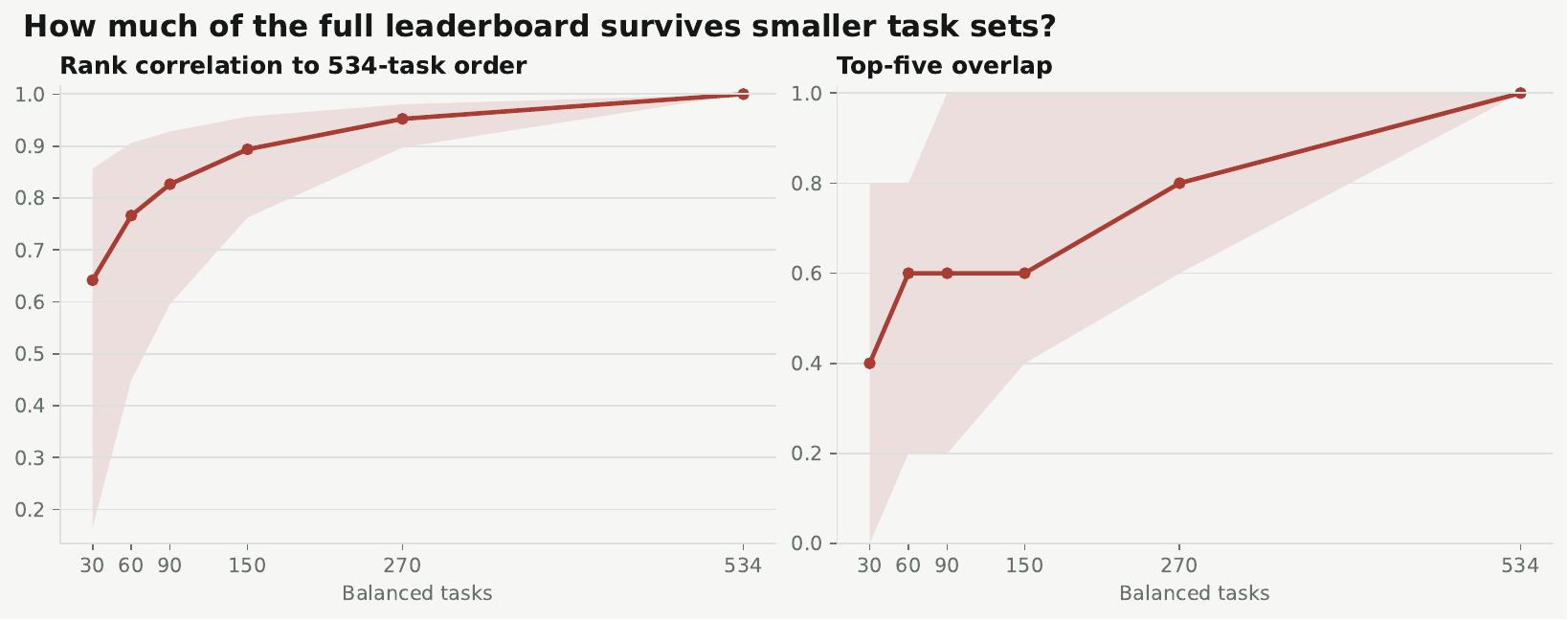}
  \caption{Stability of smaller, score-blind task subsets relative to the complete 534-task point
  order. Lines are medians across family--panel stratified samples; bands are empirical 2.5th to
  97.5th percentiles. The 534-task endpoint is identical to the reference by construction. This is
  a precision diagnostic, not a post-hoc power calculation.}
  \label{fig:task-count-stability}
\end{figure*}

\subsection{Which pairwise conclusions survive correction?}

\begin{figure}[t]
  \centering
  \includegraphics[width=\linewidth]{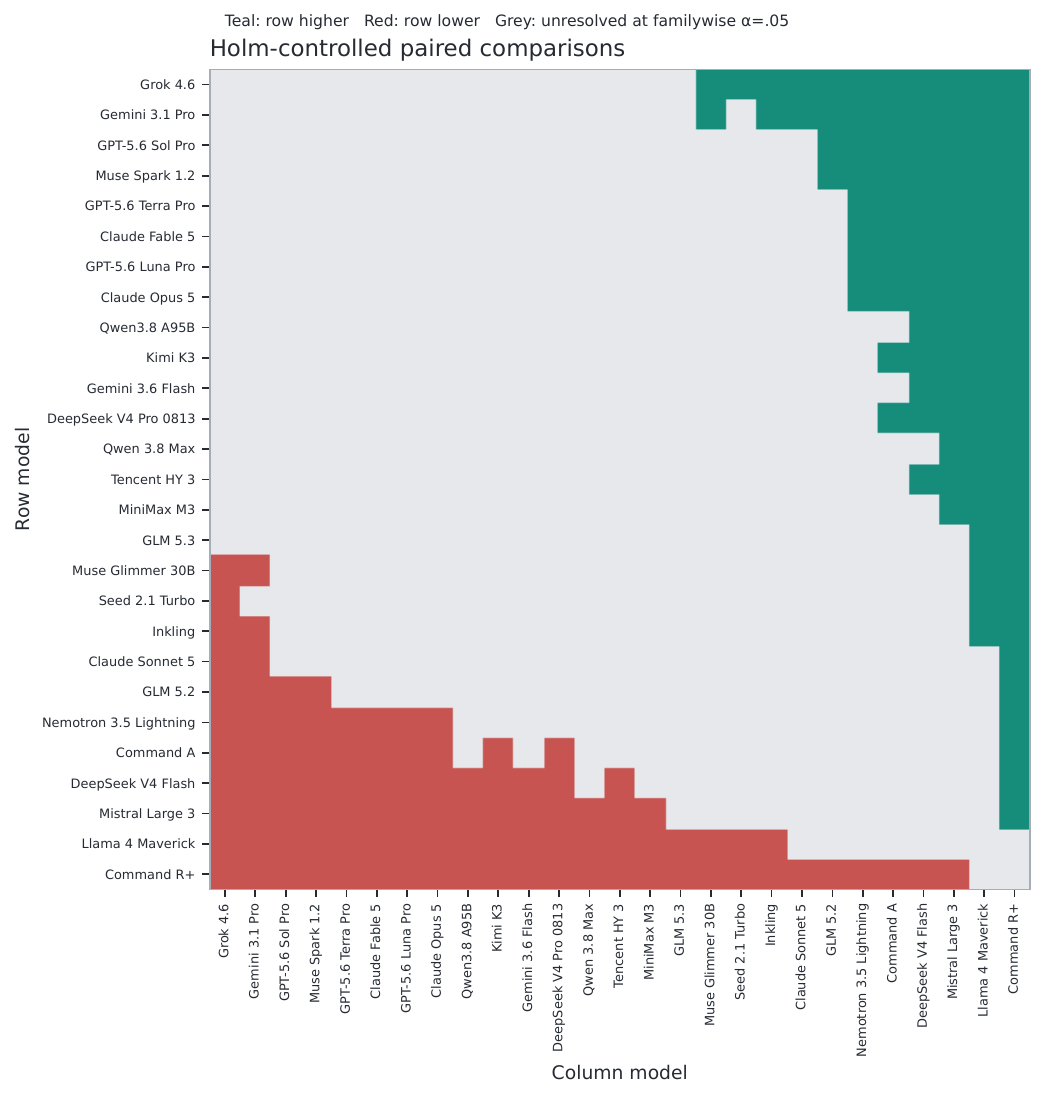}
  \caption{All paired model contrasts after Holm correction. Teal means the row model scores
  higher; red means lower; grey is unresolved.}
  \label{fig:pairwise}
\end{figure}

The score is continuous, and each model pair is compared on the same \FBTasks{} tasks rather than
on a small number of arena votes. \cref{fig:pairwise} visualizes the
multiplicity-controlled results. Grey cells are scientifically useful: they mark pairs for which
the shared task evidence does not support a directional claim at familywise $\alpha=.05$. This is
different from declaring the two models equal.

\subsection{Does the task filter or score formula determine the order?}

The endpoint whose candidate-panel validity most constrained the common-core filter was
\FBLOOModel{}. Omitting it and reapplying the original task-selection hash retains
\FBLOOOverlap{} of the official tasks. Among the remaining 26 endpoints, the new scores differ by
\FBLOOMeanShift{} points on average and at most \FBLOOMaxShift{} points; rank correlation is
\FBLOORankRho{}, \FBLOOPairAgreement{} of pair directions agree, and the point leader is
unchanged. The other omissions alter at most three tasks per stratum.

The selected and unselected candidate tasks differ by at most \FBSelectionMaxSMD{} pooled standard
deviations across the six numerical diagnostics. None of the 42 numerical or category comparisons
is resolved after Holm correction against the random-subset reference. The score itself is not
the only summary that recovers the broad order: \cref{tab:score-sensitivity} shows rank
correlations from \FBMetricMinRho{} to 0.985 and pair-order agreement of at least
\FBMetricMinAgreement{}. Across \FBWeightGridPoints{} moderate family-weight combinations, rank
correlation with equal weighting never falls below \FBWeightMinRho{}. The point leader can change
between Grok 4.6 and Gemini 3.1 Pro, which reinforces the primary analysis: the global ordering is
stable, but the data do not identify a unique best endpoint.

\begin{table}[t]
  \centering
  \scriptsize
  \setlength{\tabcolsep}{3.5pt}
  \input{generated/complete-core/complete-core-score-sensitivity-table.tex}
  \caption{Sensitivity to three alternative summaries of the same 534 decisions. Correlation and
  pair order are computed against the primary FlavourBench Score. These are descriptive
  robustness checks, not additional confirmatory rankings.}
  \label{tab:score-sensitivity}
\end{table}

\subsection{Is the ordering tied to one Epicure embedding?}

The three public checkpoints change the individual reward maps substantially. Median task-level
rank correlation with the primary map ranges from 0.660 to 0.752, and the exact optimum agrees on
27.7\% to 38.4\% of tasks. Despite those local changes, the aggregate model order is similar:
model-rank correlation ranges from \FBPublicScorerMinRho{} to \FBPublicScorerMaxRho{}, and
\FBPublicScorerMinPair{} to \FBPublicScorerMaxPair{} of all model-pair directions agree. Grok 4.6
has the largest point estimate under Cooc, Core, and Chem. In the stratified bootstrap, median
model-rank correlation ranges from 0.875 to 0.930. The same sampled point leader appears under the
primary and replacement map in only 38.9\% to 53.8\% of draws, so this result supports the broad
ordering rather than a definitive winner.

\begin{table}[t]
  \centering
  \scriptsize
  \setlength{\tabcolsep}{3.2pt}
  \input{generated/complete-core/complete-core-public-scorer-table.tex}
  \caption{Post-hoc replacement of the reward map on the fixed 534 tasks and fixed model
  selections. Task-map $\rho$ is the median correlation across each task's 56 portfolios. Model
  rank and pair order compare the aggregate leaderboard with the primary map.}
  \label{tab:public-scorer-sensitivity}
\end{table}

\subsection{Do public Epicure checkpoints recover observed substitutions?}

Exact matching maps \FBExternalSubMappedEvents{} of \FBExternalSubRawTest{} Recipe1MSubs test
events to the public 1,790-ingredient vocabulary, yielding \FBExternalSubPairs{} unique directed
pairs over \FBExternalSubSources{} source ingredients. Cooc, Core, and Chem place the observed
target at equal-source within-food-group rank percentiles 0.806, 0.800, and 0.780, respectively;
their source-clustered 95\% intervals are [0.788, 0.824], [0.781, 0.819], and [0.761, 0.798]. All
three one-sided tests reject the 0.5 chance null after Holm correction
($p_{\mathrm{Holm}}<10^{-4}$). On the \FBExternalSubNovelPairs{} directed pairs absent from the
Recipe1MSubs training split, the corresponding percentiles remain 0.754, 0.735, and 0.718.
Full-vocabulary Hit@10 is 0.133--0.172, compared with an analytic random baseline of 0.0056.

\begin{table}[t]
  \centering
  \scriptsize
  \setlength{\tabcolsep}{2.6pt}
  \input{generated/complete-core/complete-core-external-substitution-validation-table.tex}
  \caption{Label-independent validation on unique mapped Recipe1MSubs test pairs. Percentile is
  the equal-source rank of the observed target among same-food-group candidates; novel pairs are
  absent from the Recipe1MSubs training split. Hit@10 uses the full 1,789-candidate vocabulary.}
  \label{tab:external-substitution}
\end{table}

\subsection{Aggregate scores conceal different culinary profiles}

\begin{figure}[t]
  \centering
  \includegraphics[width=\linewidth]{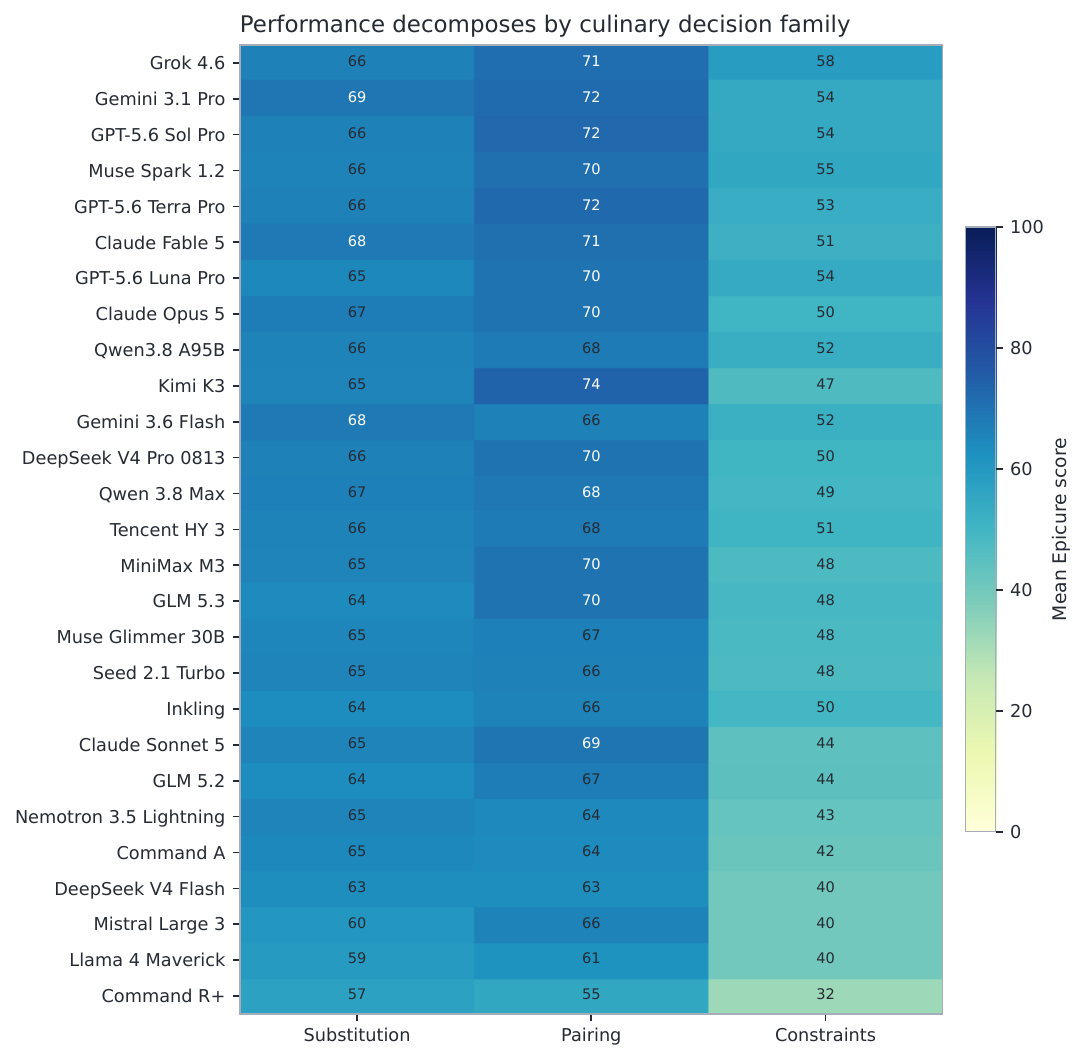}
  \caption{Mean score by family on the common core. Each family contributes
  \FBTasksPerFamily{} tasks per model. Values share the 0--100 scale but retain family-specific
  decision semantics.}
  \label{fig:families}
\end{figure}

Family-level results in \cref{fig:families} show that similar aggregate scores can arise from
different strengths. Constraint tasks penalize infeasible portfolios directly, while pairing and
substitution reward graded semantic structure. The three columns are related but not
interchangeable.

\subsection{Rank uncertainty is visible, not hidden}

\begin{figure}[t]
  \centering
  \includegraphics[width=\linewidth]{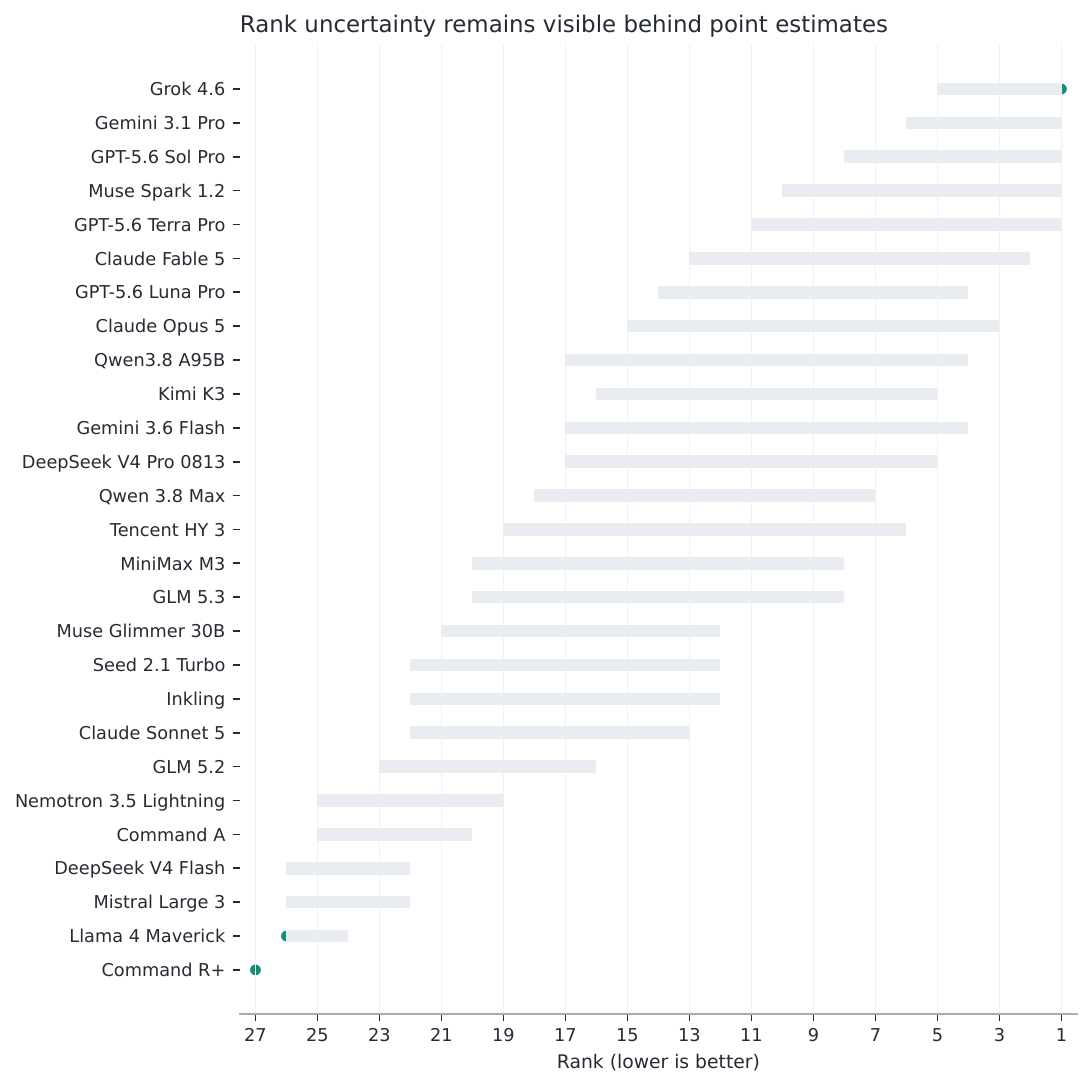}
  \caption{Point rank and 95\% bootstrap rank interval for every endpoint. Narrow intervals imply
  a stable location; overlapping intervals mark unresolved local order.}
  \label{fig:rank-intervals}
\end{figure}

Bootstrap rank intervals expose how often small score differences change order under plausible
resampling of ingredient anchors. \Cref{fig:rank-intervals} separates stable regions of the table
from visually tempting but statistically unsupported micro-ranks. This is stricter than sorting
point estimates and printing an ordinal list without uncertainty.

\subsection{Real prompts and model decisions}

\Cref{tab:examples} reports one deterministically selected common-core task from each family with
responses from the highest- and lowest-point-estimate endpoints. Selection follows a fixed hash
rule rather than task score. The public
case-study JSON contains the complete prompt, all eight choices, the
optimal portfolio, both raw model responses, selected ingredient identities, and exact scores.
The examples illustrate why continuous scoring is preferable to exact match: two non-optimal
portfolios can differ substantially in culinary utility.

\begin{table*}[t]
  \centering
  \scriptsize
  \setlength{\tabcolsep}{3pt}
  \input{generated/complete-core/complete-core-examples-table.tex}
  \caption{Three common-core response examples. Parentheses give the fixed task score. The complete
  prompts and unabridged responses are released in machine-readable form.}
  \label{tab:examples}
\end{table*}

\clearpage
\section{Reward transfer beyond format learning}
\label{sec:training}

The exhaustive maps can supervise a model as well as score it. We test whether \epicure{}
supervision transfers to unseen maps beyond learning the required answer syntax. The machine-
readable protocol and evaluator were released before optimization outcomes were generated. The
study pins Qwen3-0.6B at one repository revision \cite{yang2025qwen3}, uses 270 training and 72
validation tasks, and reserves 84 primary-transfer tasks. Ingredient anchors, candidate sets, and
prompt hashes are disjoint across those splits; all 534 leaderboard maps form a declared secondary
replication.

Three seeds receive three epochs of all-linear LoRA SFT \cite{hu2022lora} with rank 16, effective
batch size 16, learning rate $10^{-4}$, completion-only loss, and the final checkpoint without
validation-based selection. The treatment completion is each task's \epicure{} optimum. A naive
base-versus-SFT comparison would confound reward learning with practice emitting
\texttt{FINAL\_SELECTION}. We therefore rotate the same A--H portfolios onto different prompts
within each family and source panel. Treatment and control have identical prompts, row counts,
completion lengths, and portfolio-label histograms, but the control contains no accidental task
optimum. Evaluation uses greedy decoding with thinking disabled and a 64-token limit. Unparseable
completions remain in the score at zero.

The preregistered estimand is the equal-family treatment-minus-control score difference. A
50,000-draw crossed bootstrap resamples matched training seeds and ingredient anchors within the
six family--panel strata. A two-sided 100,000-draw sign-flip test operates on per-anchor effects
averaged over seeds. The interval therefore includes empirical seed and anchor variation, whereas
the $p$-value tests held-out-anchor effects conditional on the three realized seed pairs; it is not
a population-level test over retraining randomness. Three points is the preregistered practical
threshold. There is one primary contrast, so no multiplicity correction is required.

\begin{table*}[t]
  \centering
  \small
  \resizebox{0.88\textwidth}{!}{\input{generated/complete-core/complete-core-reward-transfer-table.tex}}
  \caption{Reward transfer. Scores average three training seeds except the unmodified base.
  $\Delta$ is Epicure SFT minus format control; the public-map row is a declared secondary
  replication opened only after the primary analysis was sealed.}
  \label{tab:transfer}
\end{table*}

On the 84 anchor-disjoint tasks, \epicure{} SFT scores \FBTransferPrimaryTreatment{} versus
\FBTransferPrimaryControl{} for the format control, a \FBTransferPrimaryGain{}-point gain
(95\% CI \FBTransferPrimaryCILow{}--\FBTransferPrimaryCIHigh{},
$p=\FBTransferPrimaryP{}$). All three primary seed effects are positive. Both trained arms parse
\FBTransferPrimaryControlParse{}\% of responses, while the base parses
\FBTransferPrimaryBaseParse{}\%. Format training changes the base score by only
\FBTransferPrimaryControlBaseGain{} points (95\% CI
\FBTransferPrimaryControlBaseCILow{}--\FBTransferPrimaryControlBaseCIHigh{},
$p=\FBTransferPrimaryControlBaseP{}$). The secondary public-map replication estimates a
\FBTransferPublicGain{}-point effect (95\% CI
\FBTransferPublicCILow{}--\FBTransferPublicCIHigh{}, $p=\FBTransferPublicP{}$), with all three
replication seed effects positive.

The control is empirically consequential. On the public maps it improves on the base by
\FBTransferPublicControlBaseGain{} points (95\% CI
\FBTransferPublicControlBaseCILow{}--\FBTransferPublicControlBaseCIHigh{},
$p=\FBTransferPublicControlBaseP{}$), consistent with removing parse failures. A treatment--base
comparison would fold that syntax gain into the reward effect. Treatment--control differences are
positive in all six evaluation-by-family cells and largest for constraints in both evaluations;
these family rows are descriptive rather than separately tested.

\begin{figure*}[t]
  \centering
  \includegraphics[width=0.88\textwidth]{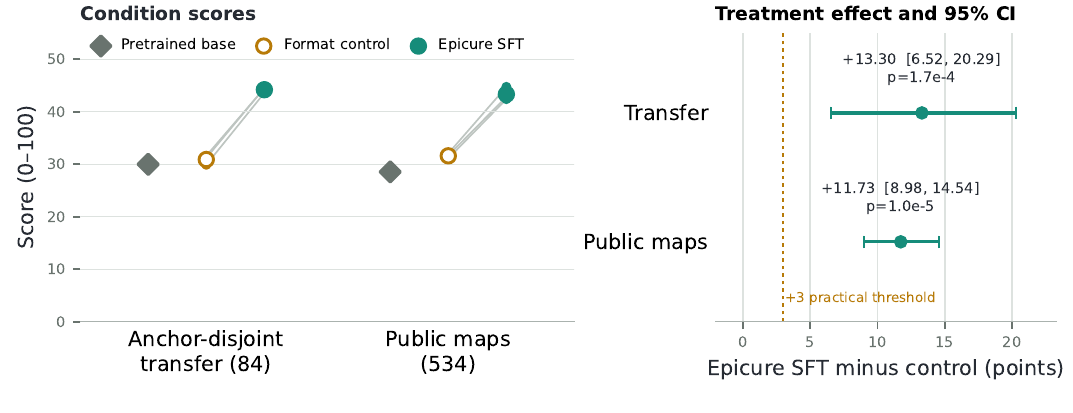}
  \caption{Held-out scores and matched treatment effects. Small points are training seeds;
  connecting lines pair control and treatment adapters initialized with the same seed. The right
  panel reports the prespecified crossed-bootstrap interval. The dashed line is the three-point
  practical threshold, not a null hypothesis.}
  \label{fig:transfer}
\end{figure*}

This is evidence for learning the released reward function rather than answer syntax. It is not
evidence of better human taste, cooked-food quality, general language-model ability,
reinforcement-learning improvement, or transfer outside this environment. The same maps expose
SFT, preference, and scalar-reward interfaces for larger follow-up studies, but only SFT is tested
here.

\clearpage
\section{Reproducibility}
\label{sec:reproducibility}

The content-addressed release
(\href{https://github.com/josefchen/flavourbench}{source},
\href{https://huggingface.co/datasets/josefchen/flavourbench}{dataset}, and
\href{https://huggingface.co/spaces/josefchen/flavourbench}{leaderboard}) includes:
\begin{itemize}
  \item both exact content-addressed task sets (\FBTasks{} scheduled tasks and
  \FBUniqueAnchors{} unique anchors), with prompts, candidates, all 56 portfolio scores, oracle
  provenance, task strata, and task-set digests;
  \item the \FBModels{}-model manifest, exact provider route per model, fallback policy, supported
  parameters, endpoint contract, and route digest;
  \item all response records used to construct the \FBCompleteCells{}-cell ranked matrix,
  including raw text, completion state, usage, latency, parser output, task score, and digest;
  \item all content-addressed provider-attempt events referenced by those responses, including
  normalized and native finish reasons used to distinguish refusals from ordinary completions;
  \item the frozen joint analysis plan, anchor-cluster contract, seeds, 50,000 bootstrap and
  100,000 sign-flip replicates, all \FBPairs{} pairwise rows, cross-panel diagnostics, and the final
  statistical release;
  \item the three pinned public-checkpoint reward maps, task-level agreement records, alternative
  leaderboards, and \FBPublicScorerReplicates{}-draw scorer-sensitivity analysis;
  \item the hash-bound Recipe1MSubs protocol, aggregate substitution-validation artifact, source-
  clustered intervals, and a builder that retrieves official raw files by immutable hash without
  redistributing them; and
  \item the frozen reward-transfer plan, format-matched control data, six training manifests,
  raw generations on both held-out evaluations, resampling outputs, and adapter hashes; and
  \item a verifier and asset builder that reconstruct the leaderboard CSV, pairwise CSV, LaTeX
  tables, and vector figures from local files without provider access.
\end{itemize}

Responses are written once under a model slot and cell identifier. Existing valid cells are
skipped on resume; conflicting artifacts fail closed. Provider fallback is disabled, so the route
shown in the manifest is the route measured. The task maps and analysis plan were frozen before
the clean primary lineage; transport-calibration responses are retained separately and never enter
the leaderboard.

\section{Related work}

General-purpose evaluation spans broad suites such as HELM \cite{liang2022helm}, difficult
knowledge tests such as MMLU-Pro and GPQA \cite{wang2024mmlupro,rein2023gpqa}, live contamination-
resistant sets \cite{white2024livebench,jain2024livecodebench}, and human preference systems such
as Chatbot Arena \cite{chiang2024chatbotarena}. Model judges increase scale but introduce their own
bias and calibration problems \cite{zheng2023judging}. \system{} instead takes the executable-
environment route used by software and agent benchmarks: SWE-bench runs repository tests
\cite{jimenez2023swebench}, BFCL evaluates function calls \cite{patil2025bfcl}, and $\tau$-bench
checks interaction outcomes against database state \cite{yao2024taubench}.
RewardBench evaluates learned reward models on chosen--rejected pairs
\cite{lambert2024rewardbench}. \system{} uses a versioned program as its reward source and asks
whether direct supervision from that program changes decisions on unseen task states after
controlling for response syntax.

Culinary datasets and benchmarks cover ingredient networks, recipe understanding, planning,
nutrition, and food knowledge \cite{ahn2011flavor,garg2018flavordb,salvador2017recipe1m,
fatemi2023substitute,choi2024cookingsense,wu2025recipe2plan,hua2025nutribench,
eftimov2026foodbench,jin2026diningbench}.
Our focus is narrower and complementary: common-task frontier model measurement against a
released culinary decision environment, with exhaustive partial-credit maps and paired inference.

\section{Limitations}

FlavourBench Score measures agreement with a released reward map, not universal human taste. The
primary runtime is reproducible by content hash, but its original training run and seed are
unrecovered; the published Epicure sibling papers do not validate that exact artifact. Rescoring
the fixed decisions with Cooc, Core, and Chem recovers the broad model order, but the original
runtime still chose the candidate sets. Recipe1MSubs supplies held-out human-observed substitution
labels for the public checkpoints, but its recipes share Recipe1M ancestry with part of Epicure's
corpus; it neither validates the unrecovered primary runtime nor tests the pairing and constraint
components. None of these checks supplies human sensory or cooking-outcome validation.

Within-task min--max normalization makes the 0--100 scale comparable while discarding absolute
raw-utility magnitudes. The tasks evaluate constrained ingredient selection rather than full recipe
generation, sensory execution, safety advice, or long-horizon kitchen planning. The common-core
requirement removes differential missingness from ranking, but it restricts the estimand to tasks
completed by every endpoint. The roster analysis measures the observed effect of this filter; it
cannot rule out dependence on task properties we did not record. Results also bind particular
model routes and collection dates. The dense reward surface supports learning experiments, but
the transfer study uses LoRA SFT of one 0.6B checkpoint over three seeds. It does not compare
training algorithms or model scales. The public-map result reuses the same adapters, so it is an
independent task replication rather than an independent training replication. Neither transfer
evaluation tests human cooking outcomes.

\section{Conclusion}

\system{} measures culinary decisions without asking one language model to grade another. A
content-addressed reward map enumerates every candidate decision, a \FBModels{}-model complete panel
produces one continuous score, and paired uncertainty distinguishes resolved differences from
apparent rank order. Controlled SFT then shows that the same maps can alter decisions on unseen
anchors beyond format practice. Under the released culinary environment, every leaderboard score,
training contrast, interval, table, and figure is reconstructable from public artifacts.

\bibliographystyle{plain}
\bibliography{references}

\clearpage
\onecolumn
\appendix
\section{Exact execution routes}
\label{app:routes}

\begin{center}
  \centering
  \scriptsize
  \setlength{\tabcolsep}{4pt}
  \input{generated/complete-core/complete-core-route-table.tex}
  \captionof{table}{Frozen execution routes by panel. Automatic fallback is disabled; route
  differences indicate complete, score-blind model-block replacements rather than cell pooling.}
\end{center}

\section{Per-family scores}

\begin{center}
  \centering
  \scriptsize
  \setlength{\tabcolsep}{4pt}
  \input{generated/complete-core/complete-core-family-table.tex}
  \captionof{table}{Mean score in each \FBTasksPerFamily{}-task family. Every entry uses the same
  tasks for every model; the equal-family mean is the FlavourBench Score.}
\end{center}

\section{Protocol constants}

\begin{center}
  \centering
  \footnotesize
  \begin{tabular}{@{}l r@{}}
    \toprule
    Quantity & Frozen value \\
    \midrule
    Models & \FBModels{} \\
    Collection panels & \FBPanelCount{} \\
    Primary tasks per model & \FBTasks{} \\
    Unique anchor clusters & \FBUniqueAnchors{} \\
    Tasks per family & \FBTasksPerFamily{} \\
    Candidate ingredients & 8 \\
    Selected ingredients & 3 \\
    Scored portfolios per task & 56 \\
    Complete response cells & \FBCompleteCells{} \\
    Bootstrap replicates & \FBBootstrapResamples{} \\
    Sign-flip replicates & \FBPermutationResamples{} \\
    Pairwise hypotheses & \FBPairs{} \\
    Shared cells per pair & \FBTasks{} \\
    Common-core rule & Status/parser only; score-blind \\
    \bottomrule
  \end{tabular}
  \captionof{table}{Core design and inference constants fixed before clean primary execution.}
\end{center}

\section{Task-selection and score sensitivity}
\label{app:robustness}

\begin{center}
  \centering
  \includegraphics[width=0.96\linewidth]{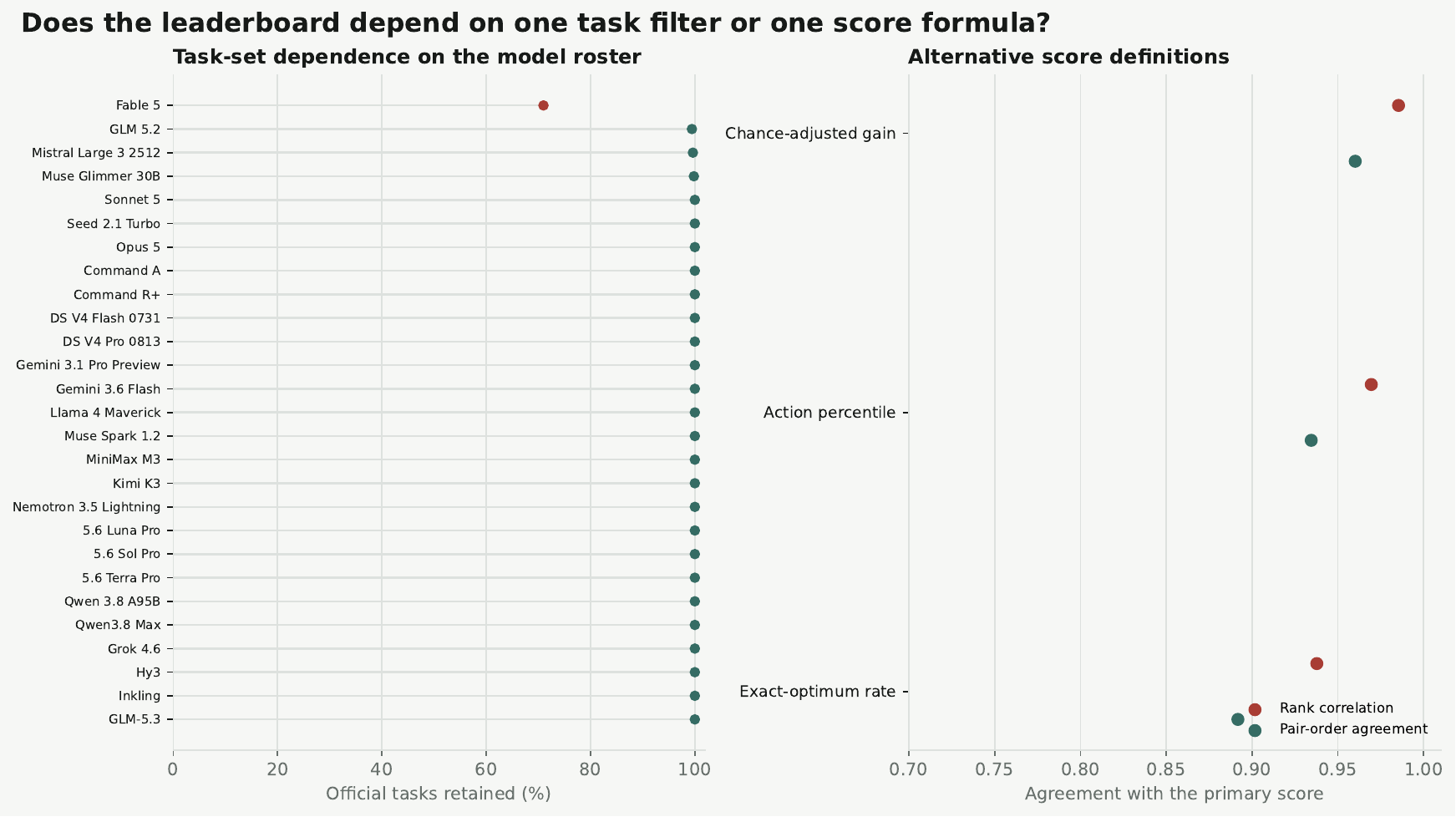}
  \captionof{figure}{Left: fraction of the official 534 tasks retained after omitting each endpoint
  and rerunning the frozen hash selection. Right: agreement between the primary score and three
  alternative summaries of the same decisions. The analysis is post-collection and does not alter
  the confirmatory release.}
\end{center}

\section{Public-checkpoint scorer sensitivity}
\label{app:public-scorer}

\begin{center}
  \centering
  \includegraphics[width=0.96\linewidth]{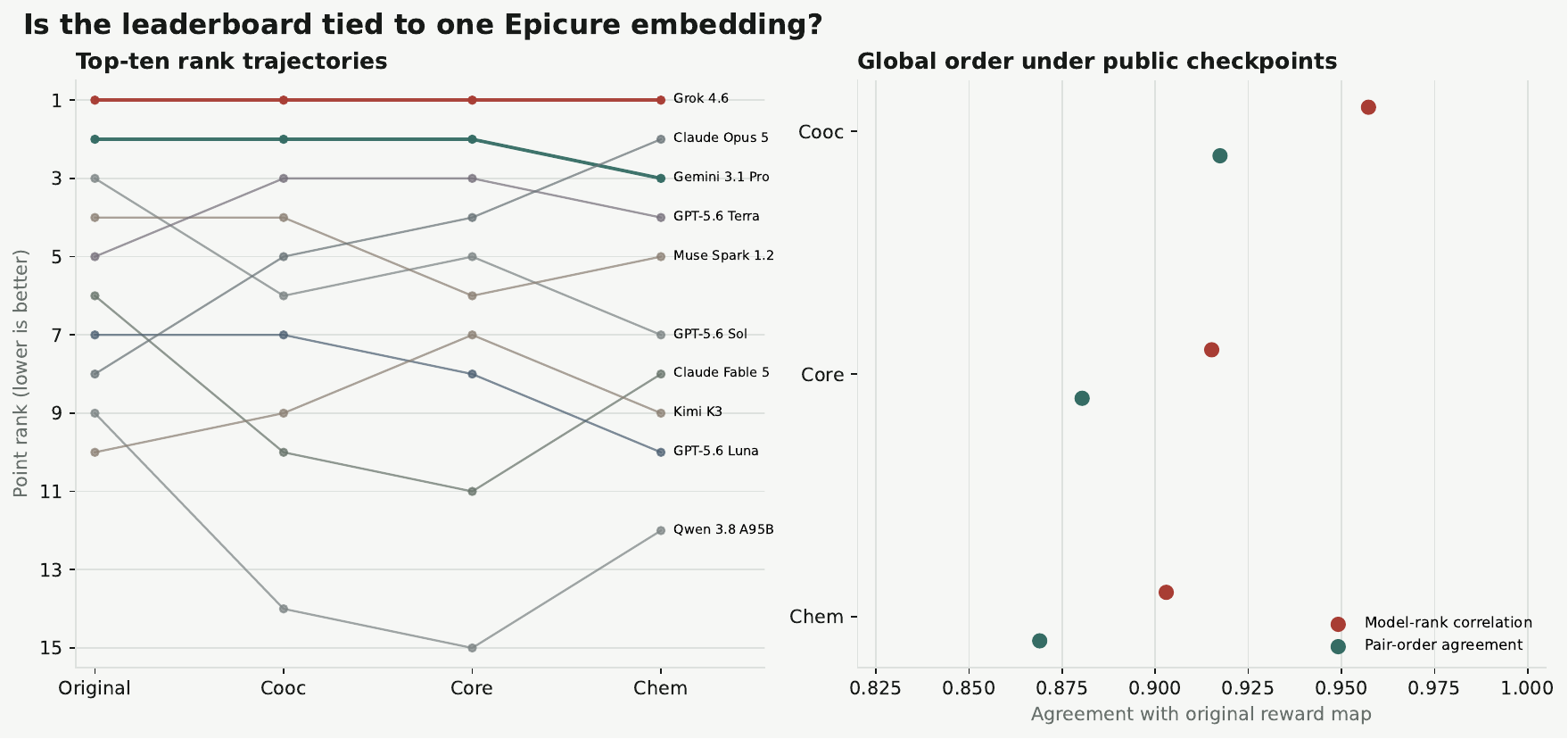}
  \captionof{figure}{Left: point-rank trajectories for the primary top ten when the same observed
  selections are scored with three public Epicure checkpoints. Right: model-rank correlation and
  model-pair direction agreement with the primary reward map. The analysis keeps the released
  tasks fixed and is post hoc.}
\end{center}

\section{External substitution validation}
\label{app:external-substitution}

\begin{center}
  \centering
  \includegraphics[width=0.96\linewidth]{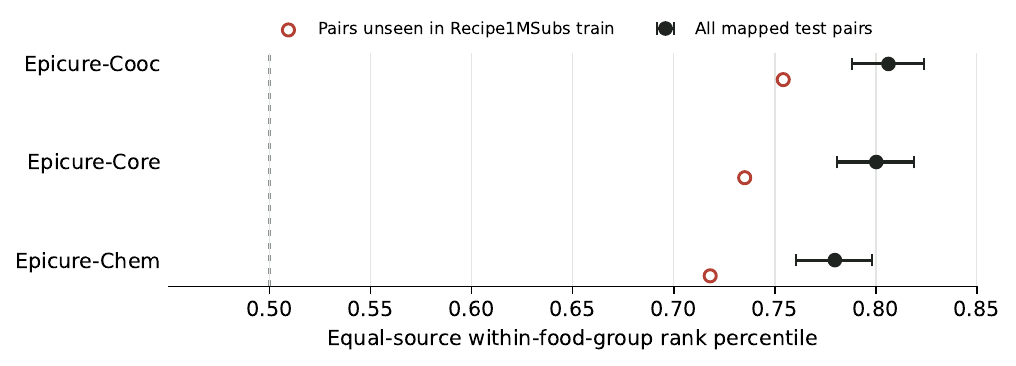}
  \captionof{figure}{Observed Recipe1MSubs targets ranked against same-food-group alternatives.
  Filled points and source-clustered 95\% intervals use all \FBExternalSubPairs{} unique mapped
  test pairs; open points retain only the \FBExternalSubNovelPairs{} directed pairs absent from
  Recipe1MSubs training. The dashed line is the 0.5 chance percentile.}
\end{center}

\section{Reward-transfer protocol and diagnostics}
\label{app:transfer}

\begin{center}
  \centering
  \small
  \begin{tabular}{@{}l l@{}}
    \toprule
    Component & Frozen value \\
    \midrule
    Base & Qwen3-0.6B, revision \texttt{c1899de289a04d1}\ldots \\
    Conditions & unmodified base; format-control SFT; Epicure-optimum SFT \\
    Train / validation / transfer & 270 / 72 / 84 anchor-disjoint tasks \\
    Training seeds & 20260824, 20260825, 20260826 \\
    LoRA & rank 16; alpha 32; dropout .05; all linear layers \\
    Optimization & 3 epochs; AdamW; LR $10^{-4}$; effective batch 16 \\
    Checkpoint rule & final adapter only; no validation selection \\
    Decoding & greedy; thinking disabled; 64 new tokens \\
    Primary score & equal family and panel; unparseable responses score zero \\
    Inference & 50,000 crossed bootstraps; 100,000 anchor sign flips \\
    \bottomrule
  \end{tabular}
  \captionof{table}{Frozen reward-transfer protocol. Validation loss is monitored but never
  selects a checkpoint.}
\end{center}

\begin{center}
  \centering
  \small
  \input{generated/complete-core/complete-core-reward-transfer-family-table.tex}
  \captionof{table}{Descriptive scores by task family. $\Delta$ is Epicure SFT minus format
  control; the family rows were not separate confirmatory hypotheses.}
\end{center}

\begin{center}
  \centering
  \small
  \input{generated/complete-core/complete-core-reward-transfer-seed-table.tex}
  \captionof{table}{Matched training-seed scores. Each $\Delta$ compares adapters initialized
  with the same seed.}
\end{center}

\end{document}

%% file: generated/complete-core/complete-core-macros.tex
\newcommand{\FBModels}{27}
\newcommand{\FBTasks}{534}
\newcommand{\FBTasksPerFamily}{178}
\newcommand{\FBTasksPerPanelFamily}{89}
\newcommand{\FBPanelCount}{2}

\newcommand{\FBUniqueAnchors}{534}
\newcommand{\FBSelectionsPerTask}{56}
\newcommand{\FBPrefrozenScores}{29904}

\newcommand{\FBPairs}{351}
\newcommand{\FBSignificantPairs}{101}
\newcommand{\FBBootstrapResamples}{50000}
\newcommand{\FBPermutationResamples}{100000}
\newcommand{\FBTopModel}{Grok 4.6}
\newcommand{\FBTopScore}{65.1}
\newcommand{\FBTopCILow}{61.0}
\newcommand{\FBTopCIHigh}{69.2}

\newcommand{\FBPanelPearson}{0.89}
\newcommand{\FBPanelSpearman}{0.80}
\newcommand{\FBIndependentClusters}{534}
\newcommand{\FBCompleteCells}{14418}

%% file: generated/complete-core/complete-core-stability-macros.tex
\newcommand{\FBStabilityReplicates}{5,000}
\newcommand{\FBGeneralizability}{0.936}
\newcommand{\FBTasksForGNinety}{329}
\newcommand{\FBHalfTaskCount}{270}
\newcommand{\FBHalfRankMedian}{0.952}
\newcommand{\FBHalfRankLow}{0.897}
\newcommand{\FBHalfRankHigh}{0.980}
\newcommand{\FBHalfTopOne}{46.1\%}
\newcommand{\FBHalfTopFive}{80\%}

%% file: generated/complete-core/complete-core-robustness-macros.tex
\newcommand{\FBLOOModel}{Claude Fable 5}
\newcommand{\FBLOOOverlap}{71.0\%}
\newcommand{\FBLOORankRho}{0.955}
\newcommand{\FBLOOPairAgreement}{92.9\%}
\newcommand{\FBLOOMeanShift}{0.65}
\newcommand{\FBLOOMaxShift}{1.37}
\newcommand{\FBSelectionMaxSMD}{0.27}
\newcommand{\FBMetricMinRho}{0.938}
\newcommand{\FBMetricMinAgreement}{89.2\%}
\newcommand{\FBWeightGridPoints}{696}
\newcommand{\FBWeightMinRho}{0.980}

%% file: generated/complete-core/complete-core-public-scorer-macros.tex
\newcommand{\FBPublicScorerMinRho}{0.903}
\newcommand{\FBPublicScorerMaxRho}{0.957}
\newcommand{\FBPublicScorerMinPair}{86.9\%}
\newcommand{\FBPublicScorerMaxPair}{91.7\%}
\newcommand{\FBPublicScorerReplicates}{20,000}

%% file: generated/complete-core/complete-core-external-substitution-validation-macros.tex
\newcommand{\FBExternalSubRawTest}{10,747}
\newcommand{\FBExternalSubMappedEvents}{3,282}
\newcommand{\FBExternalSubPairs}{1,469}
\newcommand{\FBExternalSubNovelPairs}{594}
\newcommand{\FBExternalSubSources}{357}

%% file: generated/complete-core/complete-core-reward-transfer-macros.tex
\newcommand{\FBTransferPrimaryTasks}{84}
\newcommand{\FBTransferPublicTasks}{534}

\newcommand{\FBTransferPrimaryControl}{30.90}
\newcommand{\FBTransferPrimaryTreatment}{44.20}
\newcommand{\FBTransferPrimaryBaseParse}{89.29}
\newcommand{\FBTransferPrimaryControlParse}{100.00}

\newcommand{\FBTransferPrimaryGain}{13.30}
\newcommand{\FBTransferPrimaryCILow}{6.52}
\newcommand{\FBTransferPrimaryCIHigh}{20.29}
\newcommand{\FBTransferPrimaryP}{1.70\!\times\!10^{-4}}
\newcommand{\FBTransferPrimaryControlBaseGain}{0.90}
\newcommand{\FBTransferPrimaryControlBaseCILow}{-4.91}
\newcommand{\FBTransferPrimaryControlBaseCIHigh}{6.85}
\newcommand{\FBTransferPrimaryControlBaseP}{0.790}

\newcommand{\FBTransferPublicGain}{11.73}
\newcommand{\FBTransferPublicCILow}{8.98}
\newcommand{\FBTransferPublicCIHigh}{14.54}
\newcommand{\FBTransferPublicP}{1.00\!\times\!10^{-5}}
\newcommand{\FBTransferPublicControlBaseGain}{3.05}
\newcommand{\FBTransferPublicControlBaseCILow}{0.46}
\newcommand{\FBTransferPublicControlBaseCIHigh}{5.68}
\newcommand{\FBTransferPublicControlBaseP}{0.021}

%% file: generated/complete-core/complete-core-task-table.tex
\begin{tabular}{@{}l r r r@{}}
\toprule
Family & Exact chance & Top gap & Distinct scores \\
\midrule
Substitution & 45.2 & 5.6 & 56 \\
Pairing & 45.0 & 5.6 & 56 \\
Constraints & 3.4 & 37.3 & 4 \\
\bottomrule
\end{tabular}

%% file: generated/complete-core/complete-core-leaderboard-table.tex
\begin{tabular}{@{}r l r c c c@{}}
\toprule
Rank & Model & FB Score & simultaneous 95\% CI & rank 95\% CI & group \\
\midrule
1 & Grok 4.6 & 65.1 & [61.0, 69.2] & [1, 5] & 1 \\
2 & Gemini 3.1 Pro & 65.0 & [60.8, 69.1] & [1, 6] & 1 \\
3 & GPT-5.6 Sol Pro & 64.2 & [60.1, 68.4] & [1, 8] & 1 \\
4 & Muse Spark 1.2 & 63.8 & [59.6, 67.9] & [1, 10] & 1 \\
5 & GPT-5.6 Terra Pro & 63.7 & [59.5, 67.8] & [1, 11] & 1 \\
6 & Claude Fable 5 & 63.4 & [59.2, 67.5] & [2, 13] & 1 \\
7 & GPT-5.6 Luna Pro & 62.6 & [58.5, 66.8] & [4, 14] & 1 \\
8 & Claude Opus 5 & 62.5 & [58.4, 66.6] & [3, 15] & 1 \\
9 & Qwen3.8 A95B & 62.1 & [57.9, 66.2] & [4, 17] & 1 \\
10 & Kimi K3 & 62.1 & [57.9, 66.2] & [5, 16] & 1 \\
11 & Gemini 3.6 Flash & 62.0 & [57.7, 66.2] & [4, 17] & 1 \\
12 & DeepSeek V4 Pro 0813 & 62.0 & [57.8, 66.1] & [5, 17] & 1 \\
13 & Qwen 3.8 Max & 61.5 & [57.4, 65.6] & [7, 18] & 1 \\
14 & Tencent HY 3 & 61.5 & [57.3, 65.7] & [6, 19] & 1 \\
15 & MiniMax M3 & 60.9 & [56.8, 65.1] & [8, 20] & 1 \\
16 & GLM 5.3 & 60.6 & [56.4, 64.8] & [8, 20] & 1 \\
17 & Muse Glimmer 30B & 59.9 & [55.9, 63.9] & [12, 21] & 2 \\
18 & Seed 2.1 Turbo & 59.7 & [55.4, 64.0] & [12, 22] & 2 \\
19 & Inkling & 59.6 & [55.4, 63.8] & [12, 22] & 2 \\
20 & Claude Sonnet 5 & 59.5 & [55.4, 63.7] & [13, 22] & 2 \\
21 & GLM 5.2 & 58.5 & [54.2, 62.7] & [16, 23] & 2 \\
22 & Nemotron 3.5 Lightning & 57.4 & [53.2, 61.6] & [19, 25] & 2 \\
23 & Command A & 56.7 & [52.5, 60.9] & [20, 25] & 2 \\
24 & DeepSeek V4 Flash & 55.4 & [51.1, 59.7] & [22, 26] & 2 \\
25 & Mistral Large 3 & 55.4 & [51.4, 59.5] & [22, 26] & 2 \\
26 & Llama 4 Maverick & 53.7 & [49.6, 57.7] & [24, 26] & 3 \\
27 & Command R+ & 47.9 & [43.7, 52.0] & [27, 27] & 3 \\
\bottomrule
\end{tabular}

%% file: generated/complete-core/complete-core-score-sensitivity-table.tex
\begin{tabular}{@{}l r r l@{}}
\toprule
Score summary & Rank $\rho$ & Pair order & Point leader \\
\midrule
FlavourBench Score & 1.000 & 100.0\% & Grok 4.6 \\
Chance-adjusted gain & 0.985 & 96.0\% & Gemini 3.1 Pro Preview \\
Action percentile & 0.969 & 93.4\% & Grok 4.6 \\
Exact-optimum rate & 0.938 & 89.2\% & Grok 4.6 \\
\bottomrule
\end{tabular}

%% file: generated/complete-core/complete-core-public-scorer-table.tex
\begin{tabular}{@{}l r r r l@{}}
\toprule
Reward map & Task-map $\rho$ & Model-rank $\rho$ & Pair order & Point leader \\
\midrule
Epicure-Cooc & 0.752 & 0.957 & 91.7\% & Grok 4.6 \\
Epicure-Core & 0.672 & 0.915 & 88.0\% & Grok 4.6 \\
Epicure-Chem & 0.660 & 0.903 & 86.9\% & Grok 4.6 \\
\bottomrule
\end{tabular}

%% file: generated/complete-core/complete-core-external-substitution-validation-table.tex
\begin{tabular}{lccc}
\toprule
Checkpoint & Percentile [95\% CI] & Novel pairs & Hit@10 \\
\midrule
Cooc & 0.806 [0.788, 0.824] & 0.754 & 0.133 \\
Core & 0.800 [0.781, 0.819] & 0.735 & 0.172 \\
Chem & 0.780 [0.761, 0.798] & 0.718 & 0.155 \\
\bottomrule
\end{tabular}

%% file: generated/complete-core/complete-core-examples-table.tex
\begin{tabularx}{\linewidth}{@{}l X X X@{}}
\toprule
Family & Prompt & Higher-ranked selection & Lower-ranked selection \\
\midrule
Substitution & Select three alternatives to `boursin cheese` that best preserve its dairy role, regional context, and mutual portfolio coherence. & caciocavallo, fromage blanc, grana padano (100) & fromage blanc, quail egg, goose egg (15) \\
Pairing & Select the three-ingredient bundle with the strongest learned pairing to `sweetbread` and the best internal coherence. & parsnip, cipollini onion, porcini mushroom (92) & parsnip, pickled onion, peppadew pepper (0) \\
Constraints & For a dish centered on `potato`, select three vegetarian complements with NOVA processing level at most 2; among valid portfolios, maximize learned pairing and coherence. & parsley, black pepper, thyme (100) & cheese, bay leaf, parsley (0) \\
\bottomrule
\end{tabularx}

%% file: generated/complete-core/complete-core-reward-transfer-table.tex
% Auto-generated by build_reward_transfer_assets.py.
\begin{tabular}{@{}lrrrrr@{}}
\toprule
Evaluation & Base & Format control & Epicure SFT & $\Delta$ [95\% CI] & $p$ \\
\midrule
Anchor-disjoint transfer ($n=84$) & 29.99 & 30.90 & 44.20 & +13.30 [6.52, 20.29] & $1.70\!\times\!10^{-4}$ \\
Public-map replication ($n=534$) & 28.56 & 31.60 & 43.33 & +11.73 [8.98, 14.54] & $1.00\!\times\!10^{-5}$ \\
\bottomrule
\end{tabular}

%% file: generated/complete-core/complete-core-route-table.tex
\begin{tabular}{@{}l l l l@{}}
\toprule
Model & Backend & Panel 1 route & Panel 2 route \\
\midrule
Grok 4.6 & openrouter & xai/zdr & xai/zdr \\
Gemini 3.1 Pro & openrouter & google-ai-studio & google-ai-studio \\
GPT-5.6 Sol Pro & openrouter & openai/flex & openai/flex \\
Muse Spark 1.2 & openrouter & meta & meta \\
GPT-5.6 Terra Pro & openrouter & openai/flex & openai/flex \\
Claude Fable 5 & openrouter & google-vertex/global & google-vertex/global \\
GPT-5.6 Luna Pro & openrouter & openai/flex & openai \\
Claude Opus 5 & openrouter & azure/us & azure/us \\
Qwen3.8 A95B & openrouter & alibaba & alibaba \\
Kimi K3 & openrouter & morph & morph \\
Gemini 3.6 Flash & openrouter & google-vertex/us & google-vertex/us \\
DeepSeek V4 Pro 0813 & openrouter & gmicloud/fp8 & gmicloud/fp8 \\
Qwen 3.8 Max & openrouter & alibaba & alibaba \\
Tencent HY 3 & openrouter & gmicloud/bf16 & gmicloud/bf16 \\
MiniMax M3 & openrouter & venice/fp8 & venice/fp8 \\
GLM 5.3 & zai\_coding\_direct & zai-coding-plan-direct & zai-coding-plan-direct \\
Muse Glimmer 30B & openrouter & fireworks & fireworks \\
Seed 2.1 Turbo & openrouter & seed/fp8 & seed/fp8 \\
Inkling & openrouter & deepinfra/fp8 & deepinfra/fp8 \\
Claude Sonnet 5 & openrouter & amazon-bedrock/claude-on-aws & amazon-bedrock/claude-on-aws \\
GLM 5.2 & openrouter & coreweave/fp4 & coreweave/fp4 \\
Nemotron 3.5 Lightning & openrouter & venice/fp4 & venice/fp4 \\
Command A & openrouter & cohere & cohere \\
DeepSeek V4 Flash & openrouter & deepinfra/fp4 & deepinfra/fp8 \\
Mistral Large 3 & openrouter & mistral & mistral \\
Llama 4 Maverick & openrouter & digitalocean & digitalocean \\
Command R+ & openrouter & cohere & cohere \\
\bottomrule
\end{tabular}

%% file: generated/complete-core/complete-core-family-table.tex
\begin{tabular}{@{}l r r r@{}}
\toprule
Model & Substitution & Pairing & Constraints \\
\midrule
Grok 4.6 & 66.1 & 70.9 & 58.2 \\
Gemini 3.1 Pro & 69.0 & 71.6 & 54.2 \\
GPT-5.6 Sol Pro & 66.2 & 72.5 & 54.1 \\
Muse Spark 1.2 & 65.9 & 70.3 & 55.0 \\
GPT-5.6 Terra Pro & 66.1 & 72.1 & 52.8 \\
Claude Fable 5 & 68.2 & 70.6 & 51.3 \\
GPT-5.6 Luna Pro & 64.5 & 69.7 & 53.7 \\
Claude Opus 5 & 67.2 & 69.9 & 50.4 \\
Qwen3.8 A95B & 66.0 & 67.8 & 52.5 \\
Kimi K3 & 65.5 & 73.7 & 47.0 \\
Gemini 3.6 Flash & 68.1 & 66.2 & 51.6 \\
DeepSeek V4 Pro 0813 & 66.4 & 69.5 & 49.9 \\
Qwen 3.8 Max & 66.6 & 68.5 & 49.5 \\
Tencent HY 3 & 66.0 & 67.9 & 50.5 \\
MiniMax M3 & 65.4 & 69.9 & 47.6 \\
GLM 5.3 & 63.7 & 69.6 & 48.4 \\
Muse Glimmer 30B & 65.1 & 66.8 & 47.8 \\
Seed 2.1 Turbo & 65.3 & 66.2 & 47.6 \\
Inkling & 63.6 & 65.8 & 49.5 \\
Claude Sonnet 5 & 65.3 & 69.2 & 44.1 \\
GLM 5.2 & 63.6 & 67.2 & 44.5 \\
Nemotron 3.5 Lightning & 65.3 & 64.3 & 42.5 \\
Command A & 64.7 & 63.9 & 41.6 \\
DeepSeek V4 Flash & 63.1 & 63.2 & 40.1 \\
Mistral Large 3 & 60.4 & 65.8 & 40.0 \\
Llama 4 Maverick & 59.4 & 61.4 & 40.2 \\
Command R+ & 56.7 & 54.9 & 32.0 \\
\bottomrule
\end{tabular}

%% file: generated/complete-core/complete-core-reward-transfer-family-table.tex
% Auto-generated by build_reward_transfer_assets.py.
\begin{tabular}{@{}llrrrr@{}}
\toprule
Evaluation & Family & Base & Format control & Epicure SFT & $\Delta$ \\
\midrule
Transfer & Substitution & 57.93 & 41.62 & 52.79 & +11.16 \\
Transfer & Pairing & 28.66 & 46.34 & 53.56 & +7.23 \\
Transfer & Constraint & 3.39 & 4.73 & 26.24 & +21.51 \\
\midrule
Public & Substitution & 47.63 & 47.43 & 52.39 & +4.96 \\
Public & Pairing & 34.36 & 43.59 & 53.62 & +10.02 \\
Public & Constraint & 3.69 & 3.78 & 23.99 & +20.21 \\
\bottomrule
\end{tabular}

%% file: generated/complete-core/complete-core-reward-transfer-seed-table.tex
% Auto-generated by build_reward_transfer_assets.py.
\begin{tabular}{@{}lrrrrrr@{}}
\toprule
& \multicolumn{3}{c}{Anchor-disjoint transfer} & \multicolumn{3}{c}{Public-map replication} \\
\cmidrule(lr){2-4}\cmidrule(l){5-7}
Seed & Control & Epicure & $\Delta$ & Control & Epicure & $\Delta$ \\
\midrule
20260824 & 31.61 & 44.47 & +12.86 & 32.06 & 44.74 & +12.69 \\
20260825 & 31.19 & 44.53 & +13.34 & 31.10 & 42.40 & +11.30 \\
20260826 & 29.89 & 43.59 & +13.70 & 31.65 & 42.85 & +11.20 \\
\bottomrule
\end{tabular}